%% file: main.tex
\documentclass[10pt,twocolumn,letterpaper]{article}

\ifdefined\pdfminorversion \pdfminorversion=7 \fi
\PassOptionsToPackage{dvipsnames,table}{xcolor}

\usepackage[pagenumbers]{cvpr2027/cvpr}

\input{config/packages}

\input{config/macros}

\definecolor{cvprblue}{rgb}{0.21,0.49,0.74}
\usepackage[breaklinks,colorlinks,allcolors=cvprblue]{hyperref}
\usepackage{url}

\def\paperID{*****}
\def\confName{CVPR}
\def\confYear{2027}

\title{DriftingPlanner: Drifting Generative Models for One-Step Trajectory Planning in End-to-End Autonomous Driving}

\author{Ziming Liu$^{1,*,\dagger}$\quad Yizhan Qu$^{1,2,*}$\quad Leichen Wang$^{1}$\quad Hao Sun$^{1}$\\
$^{1}$Bosch Research, Shanghai, China\quad $^{2}$Tongji University, Shanghai, China\\
{\tt\small ziming.liu@bosch.com}
}

\begin{document}

\maketitle
{\let\thefootnote\relax\footnotetext{$^{*}$Equal contribution.\quad $^{\dagger}$Corresponding author.}}

\begin{abstract}
    End-to-end planners decode the ego trajectory either by coordinate regression, which is fast but blends plausible maneuvers into one trajectory that may follow none of them, or by diffusion and flow matching, which model the trajectory distribution but need several inference steps. Drifting models move this iteration into training: a one-step generator learns to follow an attraction--repulsion field between generated and ground-truth samples. Drifting was developed for class-conditional image generation, where every class supplies many positive samples. In planning, every scene is its own condition. Pooling scenes would attract the generated trajectories of one scene to the ground truth of unrelated scenes, whereas computing the field per scene leaves a single positive, the expert trajectory. Mutual repulsion then holds the generated trajectories on a shell around it, so any single generated trajectory deviates from the ground truth. We present \textbf{DriftingPlanner}, a trajectory planner trained with the drifting loss alone, without any coordinate regression term. It computes the field per scene and removes the shell at inference: it decodes every noise vector together with its negation in one batched pass, and averaging each antithetic pair cancels the linear part of the displacement from the shell center. On NAVSIM \texttt{navtest}, DriftingPlanner reaches a state-of-the-art 93.75 PDMS, and with single-step inference its drifting decoder comes within 0.13 PDMS of the best diffusion or flow matching setting, which needs ten inference steps.
\end{abstract}

\input{sections/01-introduction}
\input{sections/02-related-work}
\input{sections/03-method}

\input{sections/04-experiments}
\input{sections/05-conclusion}

\normalem

{
    \small
    \bibliographystyle{cvpr2027/ieeenat_fullname}
    \bibliography{references/driftingplanner}
}

\end{document}

%% file: config/packages.tex
\usepackage{graphicx}%
\usepackage{amsmath}%
\usepackage{amssymb}%
\usepackage{float}
\usepackage[section]{placeins}

\usepackage{graphics}%
\usepackage{epsfig}%
\usepackage{mathptmx}%
\usepackage{times}%
\usepackage{amsmath}%
\usepackage{amssymb}%
\usepackage{pifont}%
\usepackage{algorithm}
\usepackage{algpseudocode}
\usepackage{amsmath}

\usepackage{color}
\definecolor{common}{RGB}{219, 48, 122}
\definecolor{r1}{RGB}{219, 48, 122}
\definecolor{r2}{RGB}{219, 48, 122}
\definecolor{r3}{RGB}{219, 48, 122}
\definecolor{r4}{RGB}{219, 48, 122}

\usepackage{amsmath, amssymb}
\usepackage{makecell}
\usepackage{multirow}

\usepackage{mathrsfs}
\usepackage{comment}

\usepackage{graphicx}

\usepackage[normalem]{ulem}

\usepackage[dvipsnames,table]{xcolor}
\definecolor{all}{HTML}{c32136}%
\definecolor{r1}{HTML}{ffa631}%
\definecolor{r2}{HTML}{40de5a}%
\definecolor{r3}{HTML}{1685a9}%
\definecolor{r4}{HTML}{56004f}%
\definecolor{tab}{HTML}{EEE6E9}

\usepackage{makecell}

\usepackage{graphicx}
\usepackage{float}

\newfloat{figtab}{htb}{fgtb}
\makeatletter
  \newcommand\figcaption{\def\@captype{figure}\caption}
  \newcommand\tabcaption{\def\@captype{table}\caption}
\makeatother

\usepackage{marvosym}

\usepackage{amssymb}

\usepackage{booktabs}
\usepackage{colortbl}

\definecolor{headerBlue}{RGB}{230, 242, 255}
\definecolor{headerGray}{RGB}{245, 247, 250}

\usepackage[most]{tcolorbox}%
\definecolor{softBlue}{RGB}{235, 245, 255}%
\definecolor{softGray}{RGB}{245, 247, 250}%

%% file: config/macros.tex
\usepackage{bm}
\usepackage{xspace}
\usepackage{amsbsy}
\usepackage{amssymb}
\usepackage{amsmath}
\usepackage{amsfonts}
\usepackage{euscript}
\usepackage{stmaryrd}

\usepackage{catchfile}
\newcommand{\getenv}[2][]{%
   \CatchFileEdef{\temp}{"|kpsewhich --var-value #2"}{\endlinechar=-1}
   \if\relax\detokenize{#1}\relax\temp\else\let#1\temp\fi}

\newtheorem{proposition}{\bf Proposition}

\newcommand{\bdm}{\begin{displaymath}}
\newcommand{\edm}{\end{displaymath}}

\newcommand{\method}{DriftingPlanner\xspace}
\newcommand{\sg}[1]{\llbracket #1\rrbracket}
\newcommand{\PDMS}{PDMS\xspace}
\definecolor{oursrow}{RGB}{232,244,253}

%% file: sections/01-introduction.tex
\section{Introduction}
\label{sec:intro}

End-to-end planners map raw sensor input and the ego status directly to a future ego trajectory~\cite{chitta2023transfuser,hu2023uniad,jiang2023vad,liao2025diffusiondrive,chen2024end}. Whatever the perception backbone, the trajectory that reaches the controller is produced by a trajectory decoder, and the training objective of that decoder decides which trajectory is emitted when the scene admits several.

Most planners train the decoder by coordinate regression, minimizing the $L_1$ distance between the predicted waypoints and the ground-truth trajectory~\cite{chitta2023transfuser,drivor2025}. Inference is one forward pass. The $L_1$ minimizer, however, is the per-coordinate median of the trajectories that are plausible for the scene. When a driver could either keep the lane or swerve around an obstacle, the median lies between the two, and a trajectory between two drivable maneuvers is not necessarily drivable itself. The drivable area compliance and collision metrics of a closed-loop or pseudo-simulated benchmark penalize precisely this case.

Diffusion and flow matching decoders model the trajectory distribution instead of its median~\cite{jiang2023motiondiffuser,yang2024diffusiones,liao2025diffusiondrive,zheng2024genad}. They pay for it at test time. Denoising takes several inference steps, and the strongest driving variants truncate the schedule to two steps around trajectory anchors~\cite{liao2025diffusiondrive}. Training is also rarely purely generative: auxiliary $L_1$ losses on the denoised trajectory or learned scorers that rank candidate trajectories~\cite{liao2025diffusiondrive,li2024hydramdp} decide much of the final output, so the reported gain cannot be attributed to the generative objective alone.

Drifting models~\cite{deng2026drifting} move the iteration from inference to training. A generator $g_\theta$ maps a noise vector to a trajectory in one pass. During training, a kernel field $\mathbf{V}_{p,q}$ pulls each generated trajectory toward the ground-truth trajectories and pushes it away from the other generated trajectories, and the generator regresses its output onto the stop-gradient target $\sg{\mathbf{x}+\eta\mathbf{V}_{p,q}(\mathbf{x})}$. The field vanishes when the generated and ground-truth trajectory distributions coincide, so the training dynamics evolve the distribution of generated trajectories toward that of the ground-truth trajectories while the deployed model stays a single forward pass. This raises a direct question for planning: can a trajectory decoder trained with the drifting loss alone, without any coordinate regression term, compete with regression and iterative generative decoders while keeping single-step inference?

Applied to planning without modification, the original drifting model cannot work. Deng et al.~\cite{deng2026drifting} develop it for class-conditional image generation and compute the field separately for each class label, with positives drawn from many images of that class. Planning differs in two ways.

First, planning has no discrete conditions. Every scene is its own condition, defined by its sensor input and ego status, so no other example in the data shares its target distribution. Pooling the scenes of a mini-batch would attract the generated trajectories of scene $b$ to the ground-truth trajectory of an unrelated scene $b'$ and repel them from its generated trajectories, so the field has to be computed per scene.

Second, computing the field per scene leaves a single positive, because driving datasets record exactly one ground-truth trajectory per scene. With $N$ generated trajectories and one positive, attraction pulls all of them to the same point, while repulsion pushes them apart. The generated trajectories settle on a shell around the ground-truth trajectory whose radius is set by this balance and not by any variance in the data, so a single generated trajectory at test time is a point on the shell rather than its center. When the noise branch of the generator ends with a LayerNorm, the noise feature has a fixed norm for every noise vector, so the generator cannot shrink the shell by scaling down its noise feature and must instead learn to suppress it in later layers. The shell is not the only source of spread. The generator is shared across scenes, and scenes with similar conditions have different ground-truth trajectories, so the generated trajectories also reflect this variability in the data. In the scenes we inspect, it appears as variation within a single driving decision, such as different curvatures or speed profiles of one left turn, rather than across decisions such as keeping the lane or changing lanes. A planner with a scorer can use these variations as local candidates around each proposal (Sec.~\ref{sec:noise_modes}).

We present \textbf{\method}, a trajectory planner trained with the drifting loss alone. It computes the drifting field per scene and otherwise keeps the field of the original drifting model, including its joint normalization of attraction and repulsion, cross-mass weighting, field normalization, and multiple kernel temperatures; we compute the normalization per scene. When the decoder refines many proposals, a coordinate-wise kernel weights each coordinate by its own error, so that a large longitudinal error no longer masks lateral and heading errors. At inference, \method decodes fixed antithetic pairs $\{\mathbf{z}_m,-\mathbf{z}_m\}$ in one batched decoder pass. A single-query planner averages the outputs, which cancels every odd-order term of the generator in $\mathbf{z}$, including the linear term that sets the shell radius; a planner with a scorer keeps them as candidates. A generator without LayerNorm on its noise path removes the norm constraint during training. A single trajectory query decoded with several latents already yields a set of distinct trajectory hypotheses, without anchors or a mode classifier, because the shared generator sees the different ground-truth trajectories of scenes with similar conditions.

Our contributions are:
\begin{itemize}
    \item A formulation of drifting for trajectory planning, in which every scene is its own condition with a single positive, and an analysis of the repulsion shell that this single-target setting produces. The resulting decoder is trained without any coordinate regression loss.
    \item Antithetic inference, which removes all odd-order terms of the generator output in one batched forward pass. On a single-query decoder, averaging over latents raises planning performance without retraining, and antithetic pairs make the output deterministic at the same cost.
    \item A coordinate-wise kernel for decoders that refine many trajectory proposals, which weights each coordinate by its own error and raises planning performance over the isotropic kernel.
    \item The observation that a single drifting query yields distinct trajectory hypotheses within one driving decision through noise sampling alone, without anchors or a mode classifier. We explain it by the generator being shared across scenes: the ground-truth trajectories of scenes with similar conditions jointly act as a positive set with several trajectories, even though each scene contributes only one.
    \item With single-step inference, a drifting decoder reaches essentially the same score as the best iterative diffusion or flow matching setting, which needs ten inference steps.
\end{itemize}

%% file: sections/02-related-work.tex
\section{Related Work}
\label{sec:related}

\noindent\textbf{End-to-end planning.}
End-to-end planners have moved from reactive convolutional policies to query-based transformers that share one representation between perception and planning~\cite{chen2024end}. TransFuser~\cite{chitta2023transfuser} fuses camera and LiDAR features with attention and regresses ego waypoints with an $L_1$ loss. UniAD~\cite{hu2023uniad}, VAD~\cite{jiang2023vad}, and PARA-Drive~\cite{weng2024paradrive} add detection, tracking, mapping, and motion forecasting as intermediate tasks before the planning query decodes the ego trajectory. DrivoR~\cite{drivor2025} replaces the task-specific perception stack with a DINOv2 ViT-S/14~\cite{oquab2024dinov2} adapted by LoRA~\cite{hu2022lora} and compresses each camera into a few register tokens. Across these designs the backbone has changed, but the trajectory head is still trained by pointwise regression to the ground-truth trajectory. \method keeps these backbones and replaces only the regression objective of the trajectory head.

\noindent\textbf{Generative trajectory planning.}
Generative planners model the conditional distribution of the trajectory rather than a point estimate. MotionDiffuser~\cite{jiang2023motiondiffuser} and Diffusion-ES~\cite{yang2024diffusiones} apply denoising diffusion~\cite{ho2020ddpm,song2021ddim} to motion prediction and planning, and GenAD~\cite{zheng2024genad} frames end-to-end driving as generative trajectory modeling in a latent space. DiffusionDrive~\cite{liao2025diffusiondrive} truncates the diffusion schedule around 20 k-means anchor trajectories and denoises in two steps. Flow matching~\cite{lipman2023flow} replaces the stochastic path with a straight interpolation but still integrates an ODE at test time. Two traits recur in this line of work: inference takes several network steps, and the final trajectory is shaped by auxiliary regression losses or by a learned scorer over candidate trajectories~\cite{liao2025diffusiondrive,li2024hydramdp}. \method uses single-step inference and no regression loss.

\noindent\textbf{One-step generative models and drifting.}
Several lines of work reduce generation to one or a few network evaluations. Distillation compresses a pretrained diffusion model into a one-step generator, either along its sampling trajectory~\cite{salimans2022progressive} or by matching distributions through score differences~\cite{luo2023diffinstruct,yin2024dmd,zhou2024sid}. Consistency models~\cite{song2023consistency,lu2024scm} and flow-map formulations~\cite{boffi2025flowmap,frans2024shortcut,geng2025meanflow} train one-step or few-step models from scratch by approximating the trajectories of the underlying ODE, and inductive moment matching~\cite{zhou2025imm} reaches few-step generation without an ODE. An older family trains a one-step generator directly, with a kernel two-sample discrepancy such as the maximum mean discrepancy~\cite{dziugaite2015mmd,li2015gmmn,li2017mmdgan} or a Sinkhorn divergence~\cite{genevay2018sinkhorn}, or with an adversarial discriminator~\cite{goodfellow2014gan}. In robot learning, one-step policies have been obtained by distilling diffusion policies~\cite{prasad2024consistencypolicy}. Drifting models~\cite{deng2026drifting} avoid any time-indexed path: a generator outputs the final sample directly, and a kernel attraction--repulsion field computed between generated and data samples provides its regression target. The field relates drifting to these implicit generators. With a Gaussian kernel it is a difference of smoothed scores, the update direction of distribution matching distillation~\cite{yin2024dmd}; its softmax-weighted attraction and repulsion resemble contrastive representation learning~\cite{zheng2021cacr}; and regressing onto a stop-gradient refinement of the model's own output is shared with recent one-step generators~\cite{zhou2025tvm,yu2025selfe}. Deng et al.\ study two-dimensional toy distributions and class-conditional image generation, where the field is computed per class label and every class supplies many positive samples. In trajectory planning each scene is its own condition with a single ground-truth trajectory, which is the setting in which inference from a single generated trajectory fails. \method computes the field per scene, keeps the field construction of Deng et al., and explains why inference from a single generated trajectory is biased in this single-target setting. Our inference rule is the antithetic variates estimator of Monte Carlo integration~\cite{hammersley1956antithetic}.

%% file: sections/03-method.tex
\section{Method}
\label{sec:method}

\begin{figure*}[t]
\centering
\includegraphics[width=\textwidth]{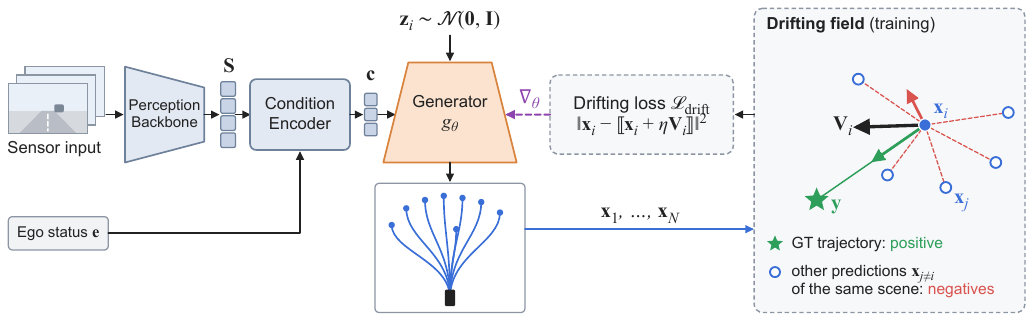}
\caption{Overview of \method. The perception backbone and the condition encoder map the sensor input and the ego status $\mathbf{e}$ to a condition $\mathbf{c}$. The generator $g_\theta$ maps $\mathbf{c}$ and a latent $\mathbf{z}_i\sim\mathcal{N}(\mathbf{0},\mathbf{I})$ to a trajectory $\mathbf{x}_i$ in a single forward pass, and $N$ latents give $N$ generated trajectories $\mathbf{x}_1,\dots,\mathbf{x}_N$ of the same scene. During training (dashed boxes), each generated trajectory $\mathbf{x}_i$ has one positive, the ground-truth (GT) trajectory $\mathbf{y}$, and $N{-}1$ negatives, the other generated trajectories $\sg{\mathbf{x}_j}$ of the same scene; generated trajectories of different scenes never form pairs. Attraction toward the positive (green arrow) and repulsion from the negatives (red arrow) sum to the drifting field $\mathbf{V}_i$ (black arrow), and $g_\theta$ is trained to move $\mathbf{x}_i$ toward the stop-gradient target $\sg{\mathbf{x}_i+\eta\mathbf{V}_i}$. The field is computed on the scaled trajectories and summed over several kernel temperatures (Sec.~\ref{sec:field}). The GT trajectory enters training only as the positive of the field; unlike regression planners, no $L_1$ or $L_2$ loss directly compares $\mathbf{x}_i$ with $\mathbf{y}$.}
\label{fig:overview}
\end{figure*}

\method keeps the perception backbone and condition encoder of an existing planner and replaces its trajectory head and training objective. We first recall drifting models (Sec.~\ref{sec:prelim}) and fix the notation (Sec.~\ref{sec:setup}). Sec.~\ref{sec:field} defines the scene-isolated drifting field, Sec.~\ref{sec:shell} analyzes the repulsion shell and introduces antithetic inference, and Sec.~\ref{sec:noise_modes} explains how a single query yields distinct trajectory hypotheses through noise sampling and how a multimodal planner with a scorer uses them. Fig.~\ref{fig:overview} summarizes the pipeline.

\subsection{Preliminaries: Drifting Models}
\label{sec:prelim}

A drifting model~\cite{deng2026drifting} trains a generator $g_\theta$ that maps a noise vector $\mathbf{z}\sim\mathcal{N}(\mathbf{0},\mathbf{I})$ to a sample $\mathbf{x}=g_\theta(\mathbf{z})$. Let $p$ denote the data distribution, which supplies the positive samples, and $q_\theta$ the distribution of generated samples, which supplies the negative samples. For a sample $\mathbf{y}$ drawn from either distribution, the kernel $k(\mathbf{x},\mathbf{y})=\exp(-\lVert\mathbf{x}-\mathbf{y}\rVert_2/\tau)$ with temperature $\tau$ measures how close $\mathbf{y}$ is to $\mathbf{x}$, and defines a field
\begin{equation}
    \mathbf{V}_{p,q}(\mathbf{x}) = \frac{\mathbb{E}_{\mathbf{y}\sim p}\big[k(\mathbf{x},\mathbf{y})(\mathbf{y}-\mathbf{x})\big]}{\mathbb{E}_{\mathbf{y}\sim p}\big[k(\mathbf{x},\mathbf{y})\big]} - \frac{\mathbb{E}_{\mathbf{y}\sim q}\big[k(\mathbf{x},\mathbf{y})(\mathbf{y}-\mathbf{x})\big]}{\mathbb{E}_{\mathbf{y}\sim q}\big[k(\mathbf{x},\mathbf{y})\big]},
    \label{eq:drift_field}
\end{equation}
whose first term attracts $\mathbf{x}$ toward the positives $\mathbf{y}\sim p$ and whose second term repels it from the negatives $\mathbf{y}\sim q$, i.e., the other generated samples. The two terms cancel when $q=p$. The generator is trained with
\begin{equation}
    \mathcal{L}(\theta) = \mathbb{E}_{\mathbf{z}}\big\lVert g_\theta(\mathbf{z}) - \sg{g_\theta(\mathbf{z})+\eta\mathbf{V}_{p,q_\theta}(g_\theta(\mathbf{z}))}\big\rVert_2^2,
    \label{eq:drift_prelim}
\end{equation}
where $\sg{\cdot}$ denotes the stop-gradient operator and $\eta>0$ is the drift step size. The gradient of~\eqref{eq:drift_prelim} moves every sample a small step along the field, so training evolves $q_\theta$ toward $p$, and inference is a single forward pass of $g_\theta$. In the original setting, $p$ is represented by many data samples. In planning, $\mathbf{x}$ and $\mathbf{y}$ are trajectories, and the field is computed within one scene, where $p$ is represented by the ground-truth trajectories of that scene. The datasets we use provide one ground-truth trajectory per scene (Sec.~\ref{sec:field}).

\subsection{Problem Setup}
\label{sec:setup}

The planner receives scene tokens $\mathbf{S}\in\mathbb{R}^{n_s\times d_s}$, e.g., camera and LiDAR tokens, and an ego status $\mathbf{e}=[\mathrm{cmd},v_x,v_y,a_x,a_y]$, where $\mathrm{cmd}$ is a one-hot driving command. It outputs a trajectory $\hat{\mathbf{O}}=(\hat{\mathbf{o}}_1,\dots,\hat{\mathbf{o}}_T)\in\mathbb{R}^{T\times 3}$ of $T$ poses, where $\hat{\mathbf{o}}_t=(o^x_t,o^y_t,\psi_t)$ holds the position in meters and the heading in radians in the current ego frame. The ground-truth trajectory of scene $b$ is $\mathbf{O}^\star_b$.

A cross-attention condition encoder maps $(\mathbf{S},\mathbf{e})$ to a condition vector $\mathbf{c}\in\mathbb{R}^{d}$. The generator $g_\theta(\mathbf{c},\mathbf{z})$ takes the condition and a latent $\mathbf{z}\sim\mathcal{N}(\mathbf{0},\mathbf{I}_{D_z})$ and predicts the trajectory directly,
\begin{equation}
    \hat{\mathbf{O}}(\mathbf{c},\mathbf{z}) = g_\theta(\mathbf{c},\mathbf{z}).
    \label{eq:param}
\end{equation}
We write $\mathbf{x}=\operatorname{vec}(\hat{\mathbf{O}})\in\mathbb{R}^D$ and $\mathbf{y}=\operatorname{vec}(\mathbf{O}^\star)$ for the generated and ground-truth trajectories, where $\operatorname{vec}$ stacks the $T$ poses into a vector of dimension $D=3T$. The ground-truth trajectory enters training only as the positive of the drifting field.

\subsection{Scene-Isolated Drifting Field}
\label{sec:field}

For every scene $b$ of a mini-batch of size $B$ we draw $N$ latents $\mathbf{z}_{b,i}$ and decode generated trajectories $\mathbf{x}_{b,i}=\operatorname{vec}(\hat{\mathbf{O}}(\mathbf{c}_b,\mathbf{z}_{b,i}))$, $i=1,\dots,N$. The positives of scene $b$ are $\mathbf{y}_{b,1},\dots,\mathbf{y}_{b,N_p}$; with one ground-truth trajectory per scene, $N_p=1$, but the formulation holds for any $N_p$. The negatives are the stop-gradient copies $\sg{\mathbf{x}_{b,j}}$ of the same scene. All kernels below are computed per scene, on a tensor of shape $B\times N\times(N_p+N)$, so no generated trajectory interacts with the targets or generated trajectories of another scene. This mirrors the per-class computation of~\cite{deng2026drifting}, with the scene in place of the class label; unlike a class, a scene supplies only its own ground-truth trajectory as positive. The rest of this section follows the field computation of~\cite{deng2026drifting}, except for the coordinate-wise kernel, the per-scene normalization, and the numerics.

\noindent\textbf{Feature scale.}
Following~\cite{deng2026drifting}, we divide all vectors by a detached scale computed over the candidate pool $\mathbf{u}_{b,1:N_p+N}=[\mathbf{y}_{b,1:N_p};\sg{\mathbf{x}_{b,1:N}}]$,
\begin{equation}
    \gamma = \max\!\Big(\frac{1}{BN(N_p+N)\sqrt{D}}\sum_{b,i,j}    \big\lVert\sg{\mathbf{x}_{b,i}}-\mathbf{u}_{b,j}\big\rVert_2,\;\gamma_{\min}\Big),
    \label{eq:scale}
\end{equation}
    with a small floor $\gamma_{\min}>0$, and set $\tilde{\mathbf{x}}=\mathbf{x}/\gamma$, $\tilde{\mathbf{y}}=\mathbf{y}/\gamma$. The scaled trajectories $\tilde{\mathbf{x}}$ keep their gradient with respect to $\theta$.

\noindent\textbf{Isotropic L2 kernel.}
For a temperature $\tau$, the kernel of Sec.~\ref{sec:prelim}, with distances divided by $\sqrt{D}$, gives the logits of scene $b$ (index $b$ dropped)
\begin{equation}
    L^{+}_{ip} = -\frac{\lVert\sg{\tilde{\mathbf{x}}_i}-\tilde{\mathbf{y}}_p\rVert_2}{\tau\sqrt{D}},\qquad
    L^{-}_{ij} = -\frac{\lVert\sg{\tilde{\mathbf{x}}_i}-\sg{\tilde{\mathbf{x}}_j}\rVert_2}{\tau\sqrt{D}},
    \label{eq:iso_kernel}
\end{equation}
with $L^{-}_{ii}=-\infty$ to exclude self-repulsion. Every pair receives one weight, shared by all $D$ coordinates. The distance is dominated by the coordinates with the largest error, in driving usually the longitudinal positions of the last poses. A generated trajectory that is far from its target along the route therefore sees nearly uniform weights over the candidate pool, and its lateral and heading errors receive only a weak, averaged correction.

\noindent\textbf{Coordinate-wise Laplace kernel.}
To decouple the coordinates, we score each coordinate $\ell=1,\dots,D$ with its own one-dimensional Laplace kernel,
\begin{equation}
    L^{+}_{ip,\ell} = -\frac{\lvert\sg{\bar{x}_{i,\ell}}-\bar{y}_{p,\ell}\rvert}{\tau},\qquad
    L^{-}_{ij,\ell} = -\frac{\lvert\sg{\bar{x}_{i,\ell}}-\sg{\bar{x}_{j,\ell}}\rvert}{\tau},
    \label{eq:coord_kernel}
\end{equation}
again with $L^{-}_{ii,\ell}=-\infty$. Here $\bar{x}_{i,\ell}=x_{i,\ell}/s_\ell$ and $\bar{y}_{p,\ell}=y_{p,\ell}/s_\ell$, and the detached scale $s_\ell$ is the standard deviation of coordinate $\ell$ over the candidate pools of the mini-batch; it replaces the shared scale $\gamma$, so every coordinate is measured in its own units. As for $\gamma$, only the scale is pooled over the mini-batch; the kernel still compares generated trajectories and targets of the same scene. Each coordinate has its own logit matrix $\mathbf{L}_\ell$ and its own weights, and the weight that corrects a lateral or heading coordinate depends only on the error along that coordinate. A large longitudinal error at the last pose no longer flattens the weights of the other coordinates. Deng et al.\ likewise sum separate drifting losses over many features of a pretrained encoder; here the features are the $D$ coordinates of the trajectory itself.

\noindent\textbf{Joint normalization.}
We concatenate $\mathbf{L}=[\mathbf{L}^{+},\mathbf{L}^{-}]\in\mathbb{R}^{N\times(N_p+N)}$ and, as in~\cite{deng2026drifting}, normalize over targets and over generated trajectories at once,
\begin{equation}
    \log\boldsymbol{\Pi} = \tfrac{1}{2}\big(\operatorname{logsoftmax}_{\mathrm{row}}(\mathbf{L})+\operatorname{logsoftmax}_{\mathrm{col}}(\mathbf{L})\big),
    \label{eq:joint_norm}
\end{equation}
which splits into an attraction block $\boldsymbol{\Pi}^{+}\in\mathbb{R}^{N\times N_p}$ and a repulsion block $\boldsymbol{\Pi}^{-}\in\mathbb{R}^{N\times N}$. Because both blocks share one normalization, a generated trajectory that is close to its target receives proportionally less repulsion, which separate softmax operators cannot express.

\noindent\textbf{Cross-mass weighting.}
Joint normalization leaves the two blocks with unequal mass, because each row splits its mass between attraction and repulsion according to the distances. Following~\cite{deng2026drifting}, each block is therefore weighted by the total mass of the other,
\begin{equation}
    W^{+}_{ip} = \Pi^{+}_{ip}\sum_{j}\Pi^{-}_{ij},\qquad
    W^{-}_{ij} = \Pi^{-}_{ij}\sum_{p}\Pi^{+}_{ip},
    \label{eq:cross_mass}
\end{equation}
so that $\sum_p W^{+}_{ip}=\sum_j W^{-}_{ij}$ for every generated trajectory $i$. The raw field is
\begin{equation}
    \mathbf{F}_i = \sum_p W^{+}_{ip}\,\tilde{\mathbf{y}}_p - \sum_j W^{-}_{ij}\,\sg{\tilde{\mathbf{x}}_j}
    = \sum_p W^{+}_{ip}(\tilde{\mathbf{y}}_p-\tilde{\mathbf{x}}_i) - \sum_j W^{-}_{ij}(\tilde{\mathbf{x}}_j-\tilde{\mathbf{x}}_i),
    \label{eq:raw_field}
\end{equation}
where the second equality uses the equal row sums. It is the discrete form of~\eqref{eq:drift_field}: attraction and repulsion are both displacements relative to $\tilde{\mathbf{x}}_i$. Without~\eqref{eq:cross_mass} the row sums differ, and $\mathbf{F}_i$ gains the term $(\sum_p W^{+}_{ip}-\sum_j W^{-}_{ij})\,\tilde{\mathbf{x}}_i$, which pushes generated trajectories away from or toward the origin of the normalized space regardless of where the target is. With equal row sums, the weights depend only on pairwise distances and the field only on relative displacements, so for a fixed feature scale, translating all generated trajectories and targets of a scene by the same vector leaves $\mathbf{F}_i$ unchanged. With the coordinate-wise kernel,~\eqref{eq:joint_norm}--\eqref{eq:raw_field} are applied to every $\mathbf{L}_\ell$ separately, with $\bar{\mathbf{x}}$ and $\bar{\mathbf{y}}$ in place of $\tilde{\mathbf{x}}$ and $\tilde{\mathbf{y}}$, and the $\ell$-th instance supplies only the component $F_{i,\ell}$ of the field.

\noindent\textbf{Numerics.}
At the sharpest temperature the weights $W^{\pm}$ underflow in single precision. We therefore keep~\eqref{eq:joint_norm}--\eqref{eq:cross_mass} in log space and subtract a per-scene maximum of the log-weights before exponentiation, which only rescales the field before the normalization below. The field is computed in \texttt{float32} with mixed precision disabled.

\noindent\textbf{Per-scene unit RMS and multiple temperatures.}
Following~\cite{deng2026drifting}, the field is normalized and summed over several kernel temperatures; we compute the normalization per scene. The field of each scene is divided by its root mean square (RMS) over the $N$ generated trajectories and $D$ coordinates, so that it has unit RMS,
\begin{equation}
    \hat{\mathbf{F}}_{b,i} = \mathbf{F}_{b,i}\Big/\sqrt{\tfrac{1}{ND}\textstyle\sum_{i'}\lVert\mathbf{F}_{b,i'}\rVert_2^2+\epsilon},
    \label{eq:unit_rms}
\end{equation}
with a small constant $\epsilon$, so every scene receives a step of the same size regardless of how far its generated trajectories are from its target. We sum the normalized fields over a set $\mathcal{T}$ of sharp and wide temperatures, $\mathbf{V}_{b,i}=\sum_{\tau\in\mathcal{T}}\hat{\mathbf{F}}^{(\tau)}_{b,i}$: the sharp kernels act on the nearest generated trajectories and the wide kernel on the layout of the whole set. The drifting loss is
\begin{equation}
    \mathcal{L}_{\mathrm{drift}} = \frac{1}{B\cdot N\cdot D}\sum_{b,i}\big\lVert\tilde{\mathbf{x}}_{b,i}-\sg{\tilde{\mathbf{x}}_{b,i}+\eta\mathbf{V}_{b,i}}\big\rVert_2^2,
    \label{eq:drift_loss}
\end{equation}
with the drift step size $\eta$ of~\eqref{eq:drift_prelim}.

\subsection{Repulsion Shell and Antithetic Inference}
\label{sec:shell}

\noindent\textbf{The repulsion shell.}
With one positive per scene, the attraction term of~\eqref{eq:raw_field} pulls all $N$ generated trajectories toward the same point $\tilde{\mathbf{y}}_b$, while the repulsion term pushes each of them away from the other $N-1$. The generated trajectories therefore settle at a finite distance around $\tilde{\mathbf{y}}_b$, a shell whose radius is set by the balance of the two terms rather than by any variability in the data. A planner, however, must emit one trajectory, and a single generated trajectory is one point on the shell instead of its center.

The architecture of the generator decides how strongly the shell persists. Consider a noise branch that ends with a LayerNorm without affine parameters, $\mathbf{h}(\mathbf{z})=\operatorname{LN}(\mathbf{W}\mathbf{z})\in\mathbb{R}^{D_h}$. Its output has zero mean and unit variance over features, so $\lVert\mathbf{h}(\mathbf{z})\rVert_2=\sqrt{D_h}$ for every $\mathbf{z}$, and $\mathbf{h}(-\mathbf{z})=-\mathbf{h}(\mathbf{z})$. The noise feature never shrinks; the generator can contract the set of generated trajectories only by learning to suppress $\mathbf{h}$ in later layers, against a repulsion that pushes them apart at every step. We call this the LN generator.

\noindent\textbf{Antithetic inference.}
The quantity a planner needs is the center of the set of generated trajectories. We estimate it with $M$ fixed antithetic pairs,
\begin{equation}
    \hat{\mathbf{x}}_M(\mathbf{c}) = \frac{1}{2M}\sum_{m=1}^{M}\big(g_\theta(\mathbf{c},\mathbf{z}_m)+g_\theta(\mathbf{c},-\mathbf{z}_m)\big),
    \label{eq:antithetic}
\end{equation}
the antithetic variates estimator of Monte Carlo integration~\cite{hammersley1956antithetic}. The following proposition states what it removes.

\begin{proposition}[Antithetic inference]
\label{prop:antithetic}
Let $g(\mathbf{c},\cdot)$ be twice continuously differentiable in $\mathbf{z}$ and write $g(\mathbf{c},\mathbf{z})=\bar{\mathbf{x}}+\mathbf{J}\mathbf{z}+\mathbf{r}(\mathbf{z})$, with $\bar{\mathbf{x}}=g(\mathbf{c},\mathbf{0})$, $\mathbf{J}=\partial g/\partial\mathbf{z}|_{\mathbf{z}=\mathbf{0}}$, and $\lVert\mathbf{r}(\mathbf{z})\rVert_2\le\frac{\beta}{2}\lVert\mathbf{z}\rVert_2^2$. Let $\mathbf{z},\mathbf{z}_1,\dots,\mathbf{z}_M\sim\mathcal{N}(\mathbf{0},\mathbf{I}_{D_z})$.
\begin{enumerate}
    \item A single generated trajectory deviates from $\bar{\mathbf{x}}$ by a root-mean-square distance of at least $\lVert\mathbf{J}\rVert_F-\frac{\beta}{2}\sqrt{D_z(D_z+2)}$.
    \item The antithetic estimator contains no term of odd order in $\mathbf{z}$: $\hat{\mathbf{x}}_M-\bar{\mathbf{x}}=\frac{1}{2M}\sum_m(\mathbf{r}(\mathbf{z}_m)+\mathbf{r}(-\mathbf{z}_m))$, and hence $\lVert\hat{\mathbf{x}}_M-\bar{\mathbf{x}}\rVert_2\le\frac{\beta}{2M}\sum_m\lVert\mathbf{z}_m\rVert_2^2$, independently of $\mathbf{J}$.
    \item $\mathbb{E}[\hat{\mathbf{x}}_M]=\mathbb{E}[g(\mathbf{c},\mathbf{z})]$, i.e., the estimator is unbiased for the mean of the generated distribution.
\end{enumerate}
\end{proposition}

\begin{proof}
(1) By the triangle inequality in $L^2$, $\sqrt{\mathbb{E}\lVert\mathbf{J}\mathbf{z}+\mathbf{r}(\mathbf{z})\rVert^2}\ge\sqrt{\mathbb{E}\lVert\mathbf{J}\mathbf{z}\rVert^2}-\sqrt{\mathbb{E}\lVert\mathbf{r}(\mathbf{z})\rVert^2}$, with $\mathbb{E}\lVert\mathbf{J}\mathbf{z}\rVert^2=\lVert\mathbf{J}\rVert_F^2$ and $\mathbb{E}\lVert\mathbf{r}(\mathbf{z})\rVert^2\le\frac{\beta^2}{4}\mathbb{E}\lVert\mathbf{z}\rVert^4=\frac{\beta^2}{4}D_z(D_z+2)$. (2) The linear terms $\mathbf{J}\mathbf{z}_m$ and $-\mathbf{J}\mathbf{z}_m$ cancel; the bound follows from the bound on $\mathbf{r}$. (3) $\mathbf{z}_m$ and $-\mathbf{z}_m$ have the same distribution as $\mathbf{z}$.
\end{proof}

The linear term $\mathbf{J}\mathbf{z}$ is the leading-order displacement of a generated trajectory from the center, i.e., the shell. Proposition~\ref{prop:antithetic} shows that a single generated trajectory carries it in full, whereas the antithetic estimator cancels it exactly for any $M\ge 1$, together with every higher odd-order term, and leaves only the even-order curvature. The $M$ pairs are drawn once and fixed, so inference is deterministic. The condition encoder runs once, and the generator processes the $2M$ inputs $(\mathbf{c},\pm\mathbf{z}_m)$ as one batch. Sec.~\ref{sec:exp_noise_modes} compares the antithetic output with individual generated trajectories.

\noindent\textbf{Direct generator.}
As an alternative to the LN generator, the direct generator is a SiLU MLP on the concatenation $[\mathbf{c};\mathbf{z}]$ without any normalization. Its first-layer weights on $\mathbf{z}$ can shrink, so the Jacobian $\mathbf{J}$ can contract during training and single generated trajectories move toward the center. Antithetic inference remains applicable and removes whatever odd-order spread is left.

\subsection{Trajectory Hypotheses from a Single Query}
\label{sec:noise_modes}

The decoder above has a single trajectory query: the condition encoder produces one vector $\mathbf{c}$ per scene, and all variation among the outputs comes from the latent $\mathbf{z}$. No anchors, mode embeddings, or mode classifier are involved. Decoding several latents for one condition nevertheless yields a set of distinct trajectory hypotheses rather than copies of one trajectory, and Sec.~\ref{sec:exp_noise_modes} measures their spread on NAVSIM.

\noindent\textbf{Where the spread comes from.}
The shell of Sec.~\ref{sec:shell} describes one scene in isolation, where a single positive leaves no data variance for the generated trajectories to represent. The generator, however, is shared across scenes and is a smooth function of $\mathbf{c}$, and scenes with similar conditions have different ground-truth trajectories. Because the target in~\eqref{eq:drift_loss} is a stop-gradient copy, the gradient with respect to a generated trajectory is the field itself up to a constant,
\begin{equation}
    \nabla_{\tilde{\mathbf{x}}_{b,i}}\mathcal{L}_{\mathrm{drift}} = -\frac{2\eta}{BND}\,\mathbf{V}_{b,i},
    \label{eq:drift_grad}
\end{equation}
so the parameter updates of scenes with nearly the same condition add. Over training, the generator at $\mathbf{c}$ is therefore driven approximately by the field averaged over the conditional distribution $p(\mathbf{y}\mid\mathbf{c})$ of ground-truth trajectories, which acts as a positive set with several ground-truth trajectories. The attraction weights $W^{+}$ decrease with the distance between a generated trajectory and a positive, so each generated trajectory is pulled mainly toward the positives near it, while the repulsion term keeps them apart. The set of generated trajectories at $\mathbf{c}$ thus reflects the variability of $p(\mathbf{y}\mid\mathbf{c})$ that the training set contains, on top of the shell. A regression decoder trained on the same scenes converges instead to a single point estimate of these trajectories, the per-coordinate median under an $L_1$ loss and the conditional mean under an $L_2$ loss. If $p(\mathbf{y}\mid\mathbf{c})$ contained separated maneuvers, the same argument would place generated trajectories on each of them; in the scenes we inspect, they spread around one maneuver (Sec.~\ref{sec:exp_noise_modes}).

\noindent\textbf{Using the hypotheses at inference.}
Antithetic inference~\eqref{eq:antithetic} returns the center of the set of generated trajectories, which is the output of a unimodal planner. When the generated trajectories spread around one maneuver, their center scores higher than a typical generated trajectory (Sec.~\ref{sec:exp_noise_modes}). A planner with a trajectory scorer can instead keep the $2M$ generated trajectories $g_\theta(\mathbf{c},\pm\mathbf{z}_m)$ as candidates. Selection, not averaging, is the right output once the generated trajectories of one condition split into separate maneuvers, because their center would then lie between them. Both outputs come from the same batched decoder pass.

\noindent\textbf{Multimodal decoder with a scorer.}
For the multimodal planner of Table~\ref{tab:main}, we follow DrivoR~\cite{drivor2025}: a set of learned trajectory queries attends to the scene tokens and is refined over several decoder stages, and each refined query serves as the condition $\mathbf{c}$ of one trajectory proposal. The drifting generator replaces the regression head that decodes each proposal. Each proposal is decoded with one antithetic pair, and a scorer head, trained on proxy \PDMS targets (Sec.~\ref{sec:exp_setup}), predicts the NAVSIM sub-scores of every candidate and selects the one with the highest predicted \PDMS.

%% file: sections/04-experiments.tex
\section{Experiments}
\label{sec:experiments}

We compare \method with published planners (Sec.~\ref{sec:exp_sota}) and then examine its design. Sec.~\ref{sec:exp_kernel} compares the isotropic and coordinate-wise kernels, and Sec.~\ref{sec:exp_temperature} varies the kernel temperatures and the drift step with paired training seeds. Sec.~\ref{sec:exp_noise_modes} measures the trajectory hypotheses of a single query, and Sec.~\ref{sec:exp_cost} compares single-step drifting inference with iterative diffusion and flow matching decoders.

\begin{table*}[t]
\centering
\caption{Comparison with state-of-the-art planners on NAVSIM \texttt{navtest}. C\&L: camera and LiDAR; C: camera only. Upper block: numbers reported in the original papers, except DrivoR$^\dagger$, whose released checkpoint we re-evaluate. Best per column in bold.}
\label{tab:main}
\resizebox{\textwidth}{!}{%
\begin{tabular}{lllcccccc}
\toprule
Method & Input & Decoder & NC $\uparrow$ & DAC $\uparrow$ & TTC $\uparrow$ & C $\uparrow$ & EP $\uparrow$ & \PDMS $\uparrow$ \\
\midrule
UniAD~\cite{hu2023uniad} & C & Regression & 97.80 & 91.90 & 92.90 & 100.00 & 78.80 & 83.40 \\
PARA-Drive~\cite{weng2024paradrive} & C & Regression & 97.90 & 92.40 & 93.00 & 99.80 & 79.30 & 84.00 \\
TransFuser~\cite{chitta2023transfuser} & C\&L & Regression & 97.70 & 92.80 & 92.80 & 100.00 & 79.20 & 84.00 \\
Hydra-MDP~\cite{li2024hydramdp} & C\&L & Scoring & 98.30 & 96.00 & 94.60 & 100.00 & 78.70 & 86.50 \\
DiffusionDrive~\cite{liao2025diffusiondrive} & C\&L & Trunc.\ diffusion & 98.20 & 96.20 & 94.70 & 100.00 & 82.20 & 88.10 \\
DrivoR$^{\dagger}$~\cite{drivor2025} & C & Regression + scorer & \textbf{98.80} & \textbf{98.95} & 96.20 & 100.00 & 90.56 & 93.68 \\
\midrule
\rowcolor{oursrow}\method & C & Drifting + scorer & 98.78 & 98.94 & \textbf{96.21} & 100.00 & \textbf{90.67} & \textbf{93.75} \\
\bottomrule
\end{tabular}}
\end{table*}

\subsection{Experimental Setup}
\label{sec:exp_setup}

\noindent\textbf{NAVSIM.}
We use the NAVSIM v1.1 splits~\cite{dauner2024navsim}, with all 103{,}288 scenes of the \texttt{navtrain} split for training and all 12{,}146 \texttt{navtest} scenes for evaluation. Each planner outputs a 4\,s trajectory of 8 poses, which the simulator tracks with an LQR controller at 10\,Hz against non-reactive traffic. We report no at-fault collision (NC), drivable area compliance (DAC), driving direction compliance (DDC), time to collision (TTC), comfort (C), ego progress (EP), and the composite
\begin{equation}
    \mathrm{PDMS} = \mathrm{NC}\times\mathrm{DAC}\times\frac{5\,\mathrm{EP}+5\,\mathrm{TTC}+2\,\mathrm{C}}{12},
    \label{eq:pdms}
\end{equation}
all scaled to $[0,100]$. DDC is reported for completeness and does not enter \PDMS. ADE and FDE (m) measure the open-loop distance to the ground-truth trajectory.

\noindent\textbf{Scorers.}
All scores reported in this paper are computed with the official NAVSIM v1 scorer, in which TTC is evaluated against 5\,s of future annotations. A GPU re-implementation of the NAVSIM sub-scores, which we call proxy \PDMS, is used only during training, where it provides the targets of the learned trajectory scorers. It evaluates the LQR rollout at 0.1\,s resolution, and its per-trajectory \PDMS has a Pearson correlation of 0.99 and a mean absolute error of 0.54 points with respect to the official scorer. It scores a batch of 4{,}096 trajectories in 31\,ms, whereas the official CPU scorer takes 117.5\,s with a single process and 1.2\,s with 32 processes.

\subsection{Comparison with the State of the Art}
\label{sec:exp_sota}

Table~\ref{tab:main} compares \method with published planners. The \method entry uses the multimodal decoder and scorer of Sec.~\ref{sec:noise_modes} on the DrivoR backbone~\cite{drivor2025}. Each of the 64 proposals is decoded with one antithetic pair, and the scorer selects one of the resulting 128 candidates; all candidates come from one batched decoder pass. \method reaches 93.75 \PDMS, above every published planner in the table.

\subsection{Kernel Design}
\label{sec:exp_kernel}

We compare the isotropic L2 kernel of Eq.~\eqref{eq:iso_kernel} with the coordinate-wise Laplace kernel of Eq.~\eqref{eq:coord_kernel}. Both planners use the multimodal decoder of Sec.~\ref{sec:noise_modes} and are trained from scratch for 25 epochs on NAVSIM \texttt{navtrain} with the drifting loss alone and seed 42. The two kernels measure distance in different units, a distance over all $D$ coordinates divided by the shared scale $\gamma$ against a one-dimensional distance in units of $s_\ell$. The comparison is therefore between the two kernels, each with its own scale and temperature set.

\begin{table}[ht]
\centering
\caption{Kernel design comparison on NAVSIM \texttt{navtest}. Both drifting decoders are trained from scratch for 25 epochs with the drifting loss alone (seed 42). Best per column in bold.}
\label{tab:kernel}
\resizebox{\columnwidth}{!}{%
\begin{tabular}{lcccccc}
\toprule
Decoder   & NC $\uparrow$ & DAC $\uparrow$ & DDC $\uparrow$ & TTC $\uparrow$ & EP $\uparrow$ & \PDMS $\uparrow$ \\
\midrule
Isotropic L2 kernel & 95.74 & 91.73 & 90.55 & 88.35 & 84.31 & 83.79 \\
\rowcolor{oursrow}Coordinate-wise Laplace kernel & \textbf{98.06} & \textbf{96.92} & \textbf{96.97} & \textbf{93.57} & \textbf{89.39} & \textbf{90.90} \\
\bottomrule
\end{tabular}}
\end{table}

Table~\ref{tab:kernel} shows that the coordinate-wise configuration raises \PDMS from 83.79 to 90.90, a gain of 7.11 points, and improves all reported sub-scores. DDC gains the most (6.42 points), followed by TTC (5.22), DAC (5.19), and EP (5.08), while NC gains 2.32. The gains in DDC and DAC are consistent with the motivation in Sec.~\ref{sec:field}: computing separate weights for each coordinate keeps the weights for lateral and heading corrections independent of longitudinal errors. These aggregate metrics do not directly establish how the correction field changes. Since the kernels also use different normalization scales and temperature sets, the 7.11-point gain reflects the full configuration change in this single-seed comparison and cannot be attributed to coordinate-wise weighting alone.

\subsection{Kernel Temperature and Drift Step}
\label{sec:exp_temperature}

\begin{figure*}[ht]
\centering
\includegraphics[width=0.5\textwidth,trim=91.2 375.1 91.7 0,clip]{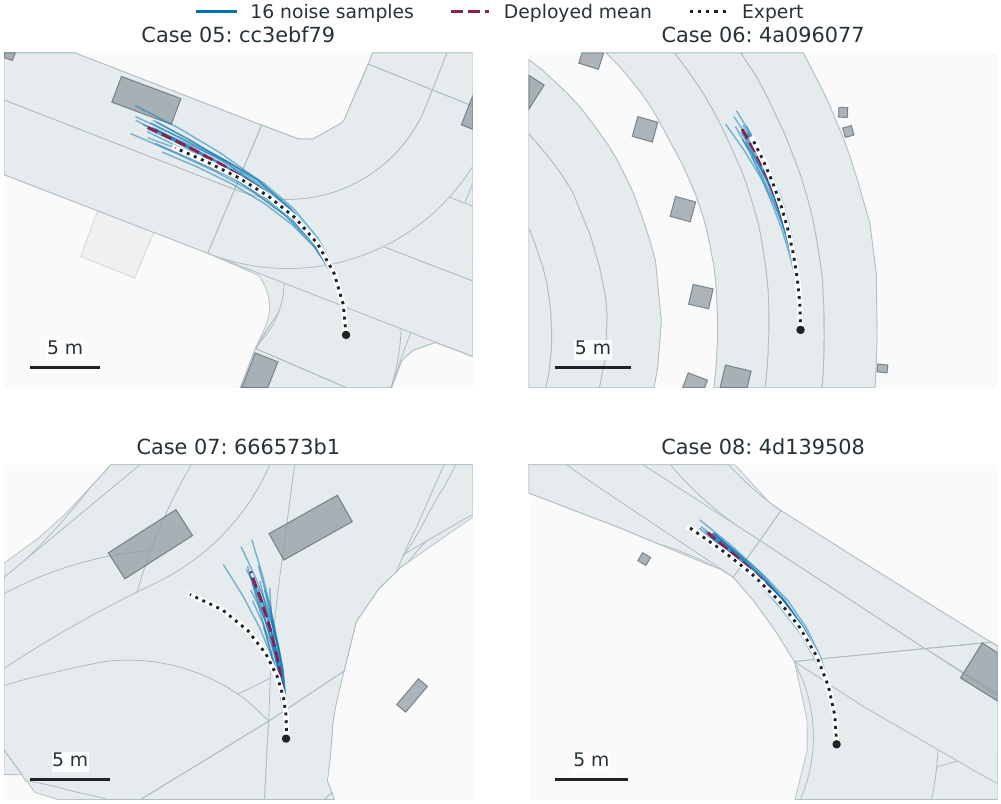}\\[2pt]
\begin{minipage}[t]{0.325\textwidth}\centering
\includegraphics[width=\linewidth,trim=1.7 199.4 253.4 25.2,clip]{figures/PUB-20261029-014_single-query-expanded-gallery/figures/gallery_left_2_double_column.pdf}\\
{\small (a) Left turn}
\end{minipage}\hfill
\begin{minipage}[t]{0.325\textwidth}\centering
\includegraphics[width=\linewidth,trim=1.7 1.7 253.4 222.5,clip]{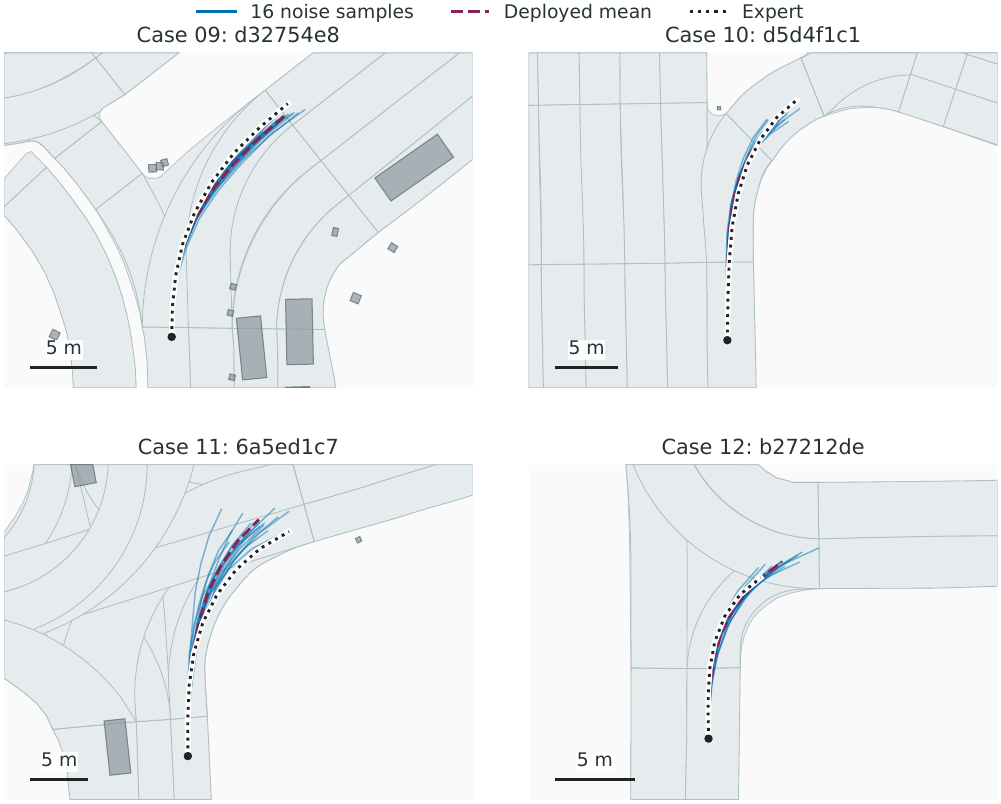}\\
{\small (b) Right turn}
\end{minipage}\hfill
\begin{minipage}[t]{0.325\textwidth}\centering
\includegraphics[width=\linewidth,trim=1.7 1.7 253.4 222.5,clip]{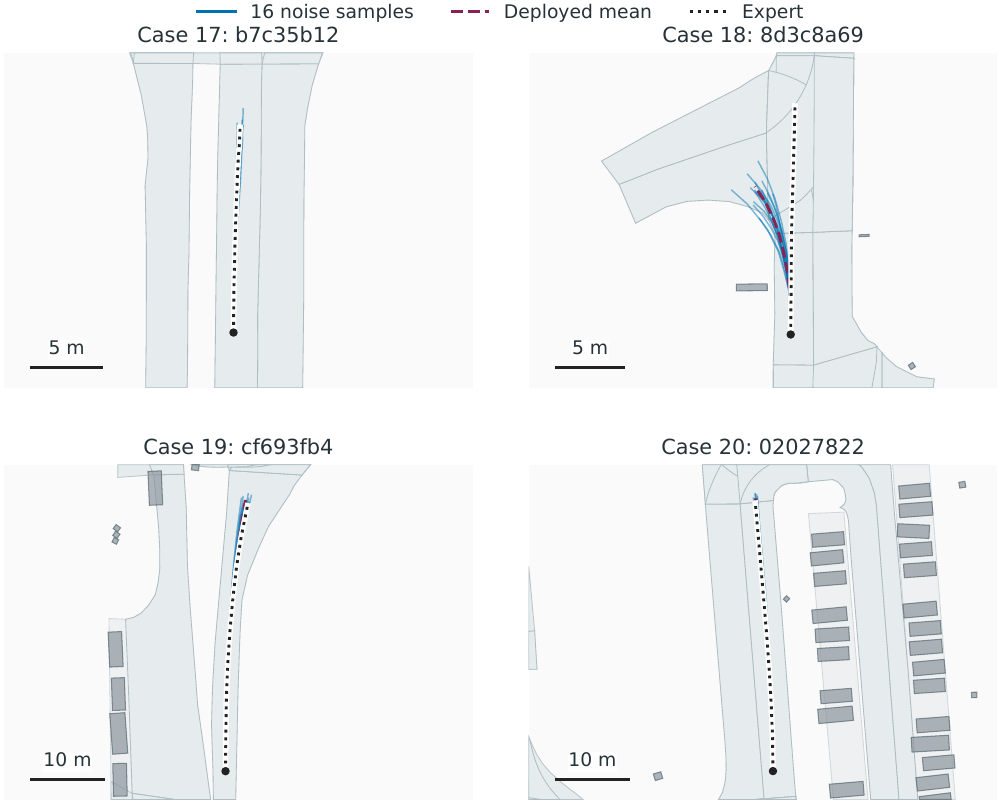}\\
{\small (c) Straight}
\end{minipage}
\caption{Generated trajectories of a single trajectory query on three NAVSIM \texttt{navtest} scenes. Blue: 16 trajectories generated from independent noise; dashed: deployed output, i.e., the antithetic mean with $M=8$; dotted: expert trajectory. (a) Left turn. (b) Right turn. (c) Straight driving.}
\label{fig:noise_modes}
\end{figure*}

We vary the two scalar hyperparameters of the field, the temperature set $\mathcal{T}$ and the drift step $\eta$, on two unimodal planners, Compact and Wide. Both take frozen ImageNet ResNet-34 tokens from the front camera (16 tokens) and LiDAR BEV tokens (16 tokens). A cross-attention encoder maps these tokens and the ego status to the condition, and an MLP generator decodes the condition together with a Gaussian latent into a residual to a constant-velocity rollout of the ego velocity. Compact uses width $d=128$ and 120 epochs, and Wide uses $d=256$ and 200 epochs. The baseline uses the isotropic kernel with joint normalization and cross-mass weighting, per-scene unit-RMS normalization, $\mathcal{T}=\{0.2,0.5,1.0\}$, and $\eta=0.1$. Training draws $N=16$ generated trajectories per scene and regresses them onto their stop-gradient drifting targets. Inference decodes one fixed latent, without antithetic averaging or candidate selection. Each variant changes one factor of the baseline, either $\mathcal{T}$ to $\{0.02,0.05,0.2\}$, the default of~\cite{deng2026drifting}, or $\eta$ to 1.0.

\begin{table}[ht]
\centering
\caption{Kernel temperature and drift step ablation study on NAVSIM \texttt{navtest}. Values are means over 8 (Compact) and 15 (Wide) paired training seeds. $\Delta$ is variant minus baseline in percentage points (pp), computed before rounding.}
\label{tab:temperature}
\resizebox{\columnwidth}{!}{%
\begin{tabular}{lrrcrr}
\toprule
 & & \multicolumn{2}{c}{$\mathcal{T}=\{0.02,0.05,0.2\}$} & \multicolumn{2}{c}{$\eta=1.0$} \\
\cmidrule(lr){3-4}\cmidrule(lr){5-6}
Track & Baseline & \PDMS & $\Delta$ (pp)  & \PDMS & $\Delta$ (pp) \\
\midrule
Compact   & 79.04 & \textbf{79.50} & $+0.46$ & 79.05 & $+0.01$ \\
Wide   & 80.18 & \textbf{80.98} & $+0.80$  & 80.35 & $+0.17$ \\
\bottomrule
\end{tabular}}
\end{table}

\noindent\textbf{Temperature.}
Table~\ref{tab:temperature} shows higher mean scores with the lower temperature set on both planners, by 0.46 points on Compact and 0.80 points on Wide. These results support the tested lower set at both feature widths, not a general claim that smaller temperatures are always better.

\noindent\textbf{Drift step.}
The step $\eta$ in Eq.~\eqref{eq:drift_loss} sets how far the stop-gradient target lies from the current generated trajectory, so a larger step might be expected to speed up training. A tenfold larger step changes \PDMS by $+0.01$ points on Compact and $+0.17$ points on Wide. This insensitivity follows from the loss. The gradient of Eq.~\eqref{eq:drift_loss} with respect to a generated trajectory is Eq.~\eqref{eq:drift_grad}, and the per-scene unit-RMS normalization fixes the magnitude of $\mathbf{V}_{b,i}$, so $\eta$ only rescales the whole gradient by a constant. AdamW divides each update by a running estimate of the gradient magnitude, which cancels such a constant up to its $\epsilon$ term, and its weight decay does not depend on the gradient. The step therefore acts neither as a learning rate nor as a regularizer here.

\subsection{Trajectory Hypotheses from a Single Query}
\label{sec:exp_noise_modes}

Sec.~\ref{sec:noise_modes} argues that one trajectory query yields distinct trajectories through the latent alone. We test this on the drifting decoder with one trajectory query trained end to end on the DrivoR backbone~\cite{drivor2025}: a DINOv2 ViT-S/14~\cite{oquab2024dinov2} with rank-32 LoRA~\cite{hu2022lora} on three cameras at $224\times336$, which yields 48 register tokens of width 384. The decoder uses the direct generator of Sec.~\ref{sec:shell}. It is trained for 15 epochs with $\mathcal{T}=\{0.02,0.05,0.2\}$ and $\eta=0.2$. On 24 fixed \texttt{navtest} scenes, eight each with a left turn, a right turn, and straight driving, we generate 16 trajectories from independent noise per scene and the antithetic output with $M=8$, and visualize them in Fig.~\ref{fig:noise_modes}. The scenes form a small predefined subset, so their scores are not comparable to full-\texttt{navtest} scores.

\begin{table}[ht]
\centering
\caption{Spread and quality of the generated trajectories of a single query on 24 fixed NAVSIM \texttt{navtest} scenes, eight per maneuver. RMS: root-mean-square distance of the 16 generated trajectories from their mean over all positions. Max pair: the largest ADE (average displacement error) between two generated trajectories of a scene. Generated: \PDMS averaged over all generated trajectories. Antithetic: Eq.~\eqref{eq:antithetic} with $M=8$. All values are averaged over scenes; the subset is not comparable to full-\texttt{navtest} scores.}
\label{tab:noise_modes}
\resizebox{\columnwidth}{!}{%
\begin{tabular}{lccccc}
\toprule
 & & \multicolumn{2}{c}{Spread (m)} & \multicolumn{2}{c}{\PDMS $\uparrow$} \\
\cmidrule(lr){3-4}\cmidrule(lr){5-6}
Maneuver & Scenes & RMS & Max pair & Generated & Antithetic \\
\midrule
Left turn & 8 & 0.50 & 1.49 & 78.30 & \textbf{80.81} \\
Right turn & 8 & 0.66 & 2.03 & 94.85 & \textbf{97.17} \\
Straight & 8 & 0.49 & 1.59 & 81.21 & \textbf{85.06} \\
\midrule
All & 24 & 0.55 & 1.70 & 84.79 & \textbf{87.68} \\
\bottomrule
\end{tabular}}
\end{table}

Fig.~\ref{fig:noise_modes} and Table~\ref{tab:noise_modes} show that varying the latent of a single query produces distinct trajectories. Their RMS distance from the per-scene mean averages 0.55\,m, while the maximum pairwise ADE averages 1.70\,m and reaches 2.74\,m. The spread increases along the prediction horizon, with an endpoint RMS of 0.98\,m. Across scenes, the RMS spread ranges from 0.26 to 0.97\,m, a nearly fourfold variation, with right turns showing the largest mean spread. This scene dependence is consistent with the influence of $p(\mathbf{y}\mid\mathbf{c})$ discussed in Sec.~\ref{sec:noise_modes}, but the experiment does not separate variability learned from the data from spread induced by the repulsion shell.

The generated trajectories vary within one maneuver rather than represent alternative maneuvers: none of the 24 scenes shows separated groups of trajectories. This diversity does not resolve errors in the predicted maneuver. In one straight-driving scene, for example, the expert continues straight, but all 16 generated trajectories turn left and score zero.

For these within-maneuver hypotheses, antithetic averaging provides a representative output. The antithetic trajectory lies 7.6\,cm from the mean of the 16 independently generated trajectories, supporting its use as an estimate of the set's center. Its mean \PDMS is 87.68, compared with 84.79 averaged over the individually generated trajectories. The antithetic score exceeds the per-scene average in 16 of the 24 scenes, ties in 7, and is lower in one.

\subsection{Iterative Decoders and Cost}
\label{sec:exp_cost}

Diffusion and flow matching iterate at inference: the decoder is applied once per denoising or integration step, so accuracy is bought with latency. Drifting performs the iteration during training, where every update moves the generated trajectories one drift step toward the stop-gradient target, and the trained generator projects noise to a trajectory in a single inference step. Table~\ref{tab:decoder_nfe} compares the two regimes on three iterative decoders: DDIM diffusion~\cite{song2021ddim}, anchor diffusion, which denoises from trajectory anchors as in DiffusionDrive~\cite{liao2025diffusiondrive}, and flow matching~\cite{lipman2023flow}. All four planners share a DINOv2 ViT-S/14 backbone initialized from pretrained weights and are otherwise trained end to end from scratch, so they differ only in the decoder and its training objective. The drifting decoder is a multimodal decoder trained with the drifting loss alone. We evaluate each iterative model with 1, 5, 10, and 20 inference steps without retraining.

\begin{table}[ht]
\centering
\caption{Drifting against iterative decoders.  Each iterative decoder is one fixed checkpoint evaluated at several numbers of function evaluations (NFE) without retraining. Latency is the median synchronized wall time at batch size 1 on a shared H20 GPU in FP32/TF32. It includes conditioning, sampling, and scoring and excludes the visual backbone and its projection. Best per column in bold.}
\label{tab:decoder_nfe}
\resizebox{\columnwidth}{!}{%
\begin{tabular}{lccccc}
\toprule
Decoder & NFE & \PDMS $\uparrow$ & ADE (m) $\downarrow$ & FDE (m) $\downarrow$ & Lat.\ (ms) $\downarrow$ \\
\midrule
\multirow{4}{*}{DDIM diffusion~\cite{song2021ddim}} & 1 & 87.61 & \textbf{0.981} & \textbf{2.329} & 6.28 \\
 & 5 & 89.02 & 1.201 & 2.834 & 7.95 \\
 & 10 & 89.65 & 1.245 & 2.949 & 9.58 \\
 & 20 & 89.83 & 1.253 & 2.973 & 13.18 \\
\midrule
\multirow{4}{*}{Anchor diffusion~\cite{liao2025diffusiondrive}} & 1 & 90.06 & 1.749 & 4.010 & 6.54 \\
 & 5 & 89.36 & 2.002 & 4.261 & 8.15 \\
 & 10 & 89.41 & 1.984 & 4.244 & 9.78 \\
 & 20 & 89.41 & 1.988 & 4.254 & 13.56 \\
\midrule
\multirow{4}{*}{Flow matching~\cite{lipman2023flow}} & 1 & 83.66 & 1.809 & 3.434 & \textbf{6.04} \\
 & 5 & 90.10 & 1.415 & 3.381 & 7.14 \\
 & 10 & \textbf{90.32} & 1.472 & 3.529 & 8.57 \\
 & 20 & 90.15 & 1.500 & 3.618 & 11.00 \\
\midrule
\rowcolor{oursrow}Drifting (ours) & 1 & 90.19 & 1.304 & 3.000 & 6.49 \\
\bottomrule
\end{tabular}}
\end{table}

With single-step inference, the drifting decoder reaches 90.19 \PDMS, above DDIM diffusion (87.61), flow matching (83.66), and anchor diffusion (90.06). DDIM and flow matching need several steps to approach this level. Flow matching gains 6.44 points from one to five steps, and DDIM gains 2.22 points from one to twenty steps but stays 0.36 points below drifting at 2.0 times its latency. Only flow matching with ten steps scores higher, by 0.13 points, at 8.57\,ms against 6.49\,ms. Anchor diffusion is the exception: its best score is at one step and drops by 0.70 points at five steps, so its multi-step rows add latency without accuracy. We attribute this to the anchors, which already place the start of the reverse process near a plausible trajectory.

Each extra step costs 0.26\,ms for flow matching and 0.36\,ms for DDIM, on top of a fixed cost of about 6\,ms that every decoder pays. With a single inference step, all four decoders run within 0.5\,ms of each other, so drifting comes within 0.13 points of the best iterative setting at the latency of a single step. The displacement metrics again do not track \PDMS. DDIM moves away from the ground-truth trajectory as it takes more steps, with ADE rising from 0.981\,m to 1.253\,m while \PDMS rises by 2.22 points. The drifting decoder has a lower ADE than every flow matching and anchor diffusion setting, but a higher ADE than every DDIM setting. We therefore read Table~\ref{tab:decoder_nfe} as evidence that single-step drifting inference matches several steps of an iterative decoder, not that it surpasses them.

%% file: sections/05-conclusion.tex
\section{Conclusion}
\label{sec:conclusion}

We presented \method, which trains an end-to-end trajectory decoder with the drifting loss alone and deploys it in a single batched forward pass. Its scene-isolated field adapts drifting to scene-specific supervision, and our analysis identifies the repulsion shell induced by a single ground-truth trajectory. Antithetic averaging cancels odd-order noise contributions for single-query inference, while a coordinate-wise kernel prevents longitudinal errors from masking lateral and heading corrections when refining multiple proposals. Experiments on NAVSIM show planning performance comparable to iterative diffusion and flow matching decoders without iterative inference. A single query also produces distinct trajectory hypotheses through noise sampling, without anchors or a mode classifier.

The observed hypotheses vary within a maneuver rather than across driving decisions. Future work will investigate multiple verified trajectories per scene to supervise distinct maneuvers directly, with candidate selection to avoid averaging across incompatible alternatives. Beyond the current non-reactive evaluation, reactive closed-loop tests are needed to assess how the decoder handles reciprocal interactions with other road users and errors that accumulate over successive predictions.